\documentclass[journal]{IEEEtran}

\usepackage[T1]{fontenc}
\usepackage{times} 
\usepackage{dcolumn}
\usepackage{endnotes}
\usepackage{graphics}
\usepackage{subfigure}
\usepackage{times}
\usepackage{epsfig}
\usepackage{graphicx} 
\usepackage{textcomp}
\usepackage{amssymb,amsmath}
\usepackage{balance}
\usepackage{url}
\usepackage{listings}
\usepackage[]{algorithm2e}
\usepackage{flushend}

\begin{document}
\title{Holographic Graph Neuron: a Bio-Inspired Architecture for Pattern
Processing}

\author{Denis~Kleyko, Evgeny~Osipov, Alexander Senior, Asad I. Khan, and Y. Ahmet \c{S}ekercio\u{g}lu % <-this % stops a space

\thanks{This work was initially presented at IEEE International Conference on Computer and Information Sciences, ICCOINS 2014 \cite{ICCOINS}. } 
\thanks{D. Kleyko and E. Osipov are with the Department of Computer  Science Electrical and Space Engineering, Lule\aa{} University of Technology, 971 87 Lule\aa{}, Sweden (e-mail: Evgeny.Osipov@ltu.se; Denis.Kleyko@ltu.se).}% <-this % stops a space
\thanks{A.I. Khan is with Clayton School of Information Technology, Monash University, Wellington Rd, Clayton, Victoria 3800, Australia (e-mail: Asad.Khan@monash.edu).}% <-this % stops a space
\thanks{A. Senior and Y. A. \c{S}ekercio\u{g}lu are with the Department of Electrical and Computer Systems Engineering, Monash University, Clayton, Victoria 3800, Australia (e-mail: Alexander.Senior@monash.edu; Ahmet.Sekercioglu@monash.edu).} }% <-this % stops a space

 \maketitle

\begin{abstract} 
\boldmath This article proposes the use of Vector Symbolic Architectures
for implementing Hierarchical Graph Neuron, an architecture for memorizing 
patterns of generic sensor stimuli. The adoption of a Vector Symbolic representation ensures a one-layered design for the approach, while
maintaining the previously reported properties and performance  characteristics  of
Hierarchical Graph Neuron, and also improving the noise  resistance of the architecture. 
The proposed architecture enables a linear (with respect to the number of stored entries)
time search for an arbitrary sub-pattern.

\end{abstract}


\section{Introduction}
\label{sect:intro}

\IEEEPARstart{G}{raph} Neuron (GN) is an approach for memorizing
patterns of generic sensor stimuli for later template matching. It is
based on the hypothesis that a better associative memory resource can
be created by changing the emphasis from high speed sequential CPU
processing to parallel network-centric processing
\cite{10_NasutionKhan,12_KhanAmin}. In contrast to
contemporary machine learning approaches, GN allows introduction of
new patterns in the learning set without the need for retraining.
Whilst doing so, it exhibits a high level of scalability i.e. its
performance and accuracy do not degrade as the number of stored
patterns increases over time.

Vector Symbolic Architectures (VSA) \cite{Kanerva:Hyper_dym11} are a
bio-inspired method of representing concepts and their meaning for
modeling cognitive reasoning. It exhibits a set of unique properties
which make it suitable for implementation of artificial general
intelligence \cite{LevyGaylor, PlateTr, Gallant}, and so, creation of
complex systems for sensing and pattern recognition without reliance
on complex computation.  In the biological world, extremely
successful applications of these approaches can be found. One example
is the ordinary house fly. A house fly is capable of conducting very
complex maneuvers albeit it possesses very little computational
capacity\footnote{In \cite{Zbikowski2005}, a house fly's properties
  are compared and contrasted with an advanced fighter plane as
  follows: ``Whereas the F-35 Joint Strike Fighter, the most advanced
  fighter plane in the world, takes a few measurements - airspeed,
  rate of climb, rotations, and so on and then plugs them into complex
  equations, which it must solve in real time, the fly relies on many
  measurements from a variety of sensors but does relatively little
  computation.''}. Another interesting biological example is the
compound eyes of arthropods. These compound eyes consist of large
number of sensors with limited and localized processing capabilities
for performing relatively complex sensing tasks \cite{Song2013}.

This article presents contributions in two domains: organization of
associative memory and properties of connectionist distributed
representation. In the first area the article introduces a novel
bio-inspired architecture, called Holographic Graph Neuron (HoloGN),
for one-shot pattern learning, which is build upon Graph Neuron's
flexible input encoding abstraction and strong reasoning capabilities
of the VSA representation. In the second area this article extends
understanding of the performance proprieties of distributed
representation, which opens a way for new applications.

The article is structured as follows. Section \ref{sect:rel_work}
presents an overview of the work related to the matter presented in
this article. The background information on the theories, concepts and
approaches used in HoloGN is described in Section
\ref{sect:background}. Sections \ref{sect:hologen} through
\ref{sect:vectorperf} present the main contribution of this article --
the design of the HoloGN architecture and its performance
characteristics.  The
conclusions are presented in Section \ref{sect:conclusions}.

\section{Related Work}
\label{sect:rel_work}

Associative memory (AM) is a sub-domain of artificial neural networks, which
utilises the benefits of content-addressable memory (CAM) \cite{1_Schultz} in microcomputers. The AM
concept was originally developed in an effort to utilise the power and speed of
existing computer systems for solving large-scale and computationally intensive problems 
by simulating biological neurosystems.

The Hierarchical Graph Neuron (HGN) approach \cite{10_NasutionKhan} is a type
of associative memory which signifies the hierarchical structure in its implementation.
Hierarchical structures in associative memory models are of interest as these have been shown to 
improve the rate of recall in pattern recognition applications. The distributed HGN scheme also allows
for better control of the network resources. This scheme compares well with contemporary approaches such 
as Self-Organizing Map and Support Vector Machine in terms of  speed and
accuracy. 

Vector Symbolic Architectures \cite{LevyGaylor} are a class of
connectionist models that use hyper-dimensional vectors (i.e.\ vectors of several
thousand elements) to encode structured information as {\em distributed} or
{\em holographic} representation. In this technique structured data
is represented by performing basic arithmetic operations on field-value
tuples. Distributed representations of data structures are an approach
actively used in the area of cognitive computing  for representing and reasoning upon semantically 
bound information \cite{Kanerva:Hyper_dym11, Kanerva:Hyper_dym13}. 

In \cite{AAAIWLevyBajracharyaGaylor} a VSA-based knowledge-representation
architecture is proposed for learning arbitrarily complex, hierarchical,
symbolic relationships (patterns) between sensors and actuators in robotics.
Recently the theory of hyper-dimensional computing, and VSA in particular, were
adopted for implementing novel communication protocols and architectures for collective communications in
machine-to-machine communication scenarios \cite{MACOM, Jakimovski}. The first
work demonstrates unique reliability and timing properties essential in the context of industrial machine-to-machine
communications. The latter work shows the feasibility of implementing collective
communications using current radio technology. This article presents an
algorithmic ground for further design of the distributed HoloGN 
on top of the architecture presented in \cite{MACOM}. 

\section{Overview of essential parts of relevant concepts and theories}
\label{sect:background}

\subsection{Hierarchical Graph Neuron} 

\begin{figure}[t!]
\centering
\includegraphics[width=1.0\columnwidth]{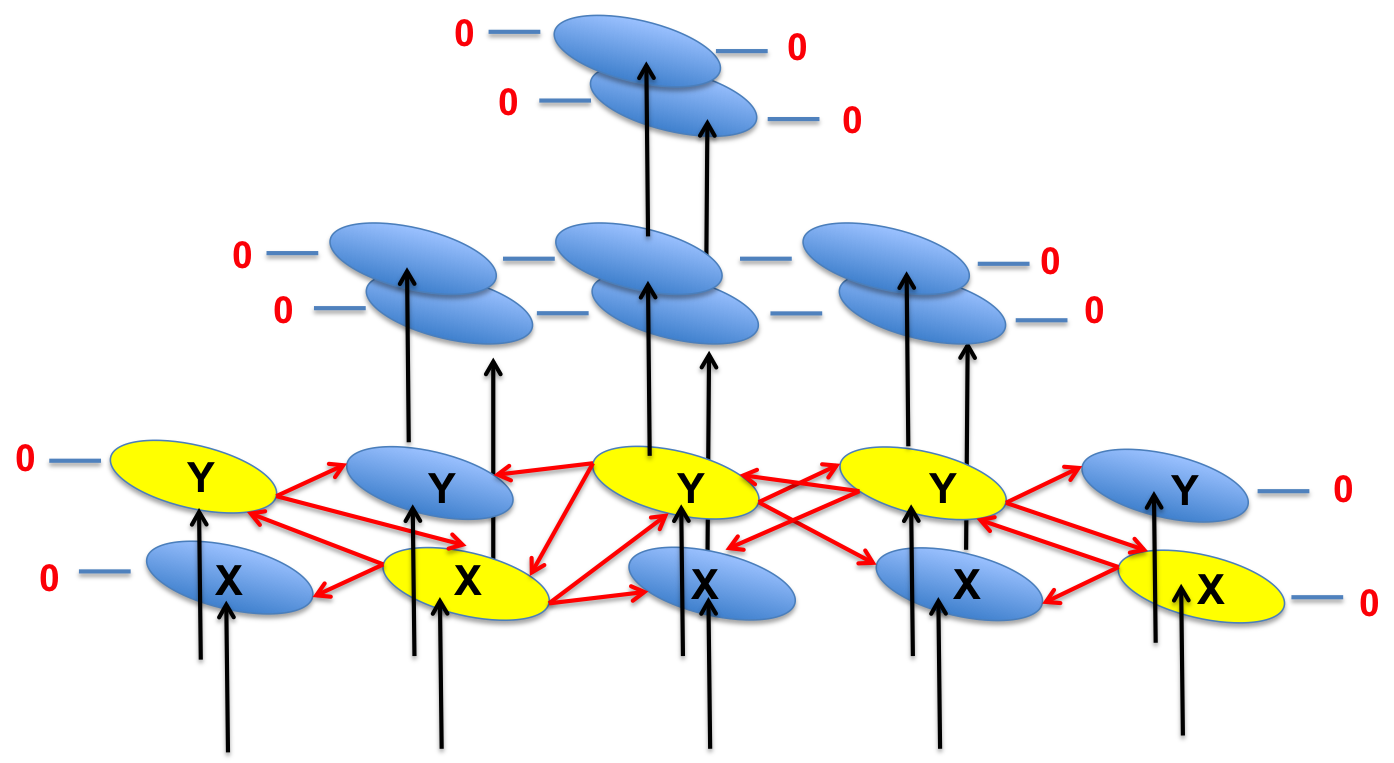}
\caption{Hierarchical GNs with five elements' pattern of two symbols. The short
arrows in the figure show how GNs communicate indices of the activated
GN-elements with their neighbours creating logical connectivity. Null GN numbers
are assigned at the start and at the end of the array for maintaining a consistent reporting scheme where every activated GN reports 
to both the adjacent GNs.}
\label{fig:HGN_1}
\end{figure}

Figure \ref{fig:HGN_1} illustrates the Hierarchical Graph Neuron approach.
Consider only the bottom layer of the construction without the hierarchy of
upper nodes; this bottom level is the original {\em flat} network of Graph
Neurons \cite{10_NasutionKhan}.
Each GN is a model for a set of generic sensory values. When seen
as a network, graph neurons can be modeled by an array where columns are
individual GNs and rows are possible symbols, which a neuron can recognize. For
example, if there are only two possible symbols, say ``X'' and ``Y'', in the
alphabet of a pattern, then only two rows are needed to represent those
symbols.
The number of columns\footnote{In this articles words ``column'' and GN are used
interchangeably and refer to a single Graph Neuron. Term ``GN array'' refers to
several GNs used to recognize a pattern of several elements, where one neuron
is used to recognize one element of the pattern.} in the GN array determines
the size of patterns, which it can analyse.

An input pattern is defined as a stimulus produced within the network. 
In Figure \ref{fig:HGN_1}, each GN can analyse a symbol (``X'' or ``Y'') of
a pattern comprising of five elements. 
In each GN (column) only the element with matching value (a row ID) would
respond.
For example, if the pattern is ``YXYYX'', then in the second column ``X''
element will be activated as response to this input stimulus. If a particular element
in the GN-column  is activated it sends a report to all 
adjacent GNs. The report contains the activated GN's element ID
(the row index).  Otherwise, it simply ignores the stimulus and returns to the idle state.

During the next phase, all GNs communicate the indices of the activated elements with the adjacent columns at their level, and additionally communicate the
stored bias information to the layer above. The procedure continues in an ascending
manner until the collective bias information reaches the top of the hierarchy. 
The higher level GNs can thus provide a more authoritative assessment of the
input pattern. The accuracy of HGN was demonstrated to be comparable to the accuracy of Neural Network with back-propagation \cite{10_NasutionKhan}.

\subsection{Fundamentals of Vector Symbolic Architecture and Binary Spatter Codes}
\label{sect:hologr_struct}

Vector Symbolic Architecture is an approach for encoding and operations on distributed representation of information. VSA has previously been used mainly in the area of cognitive computing for representing and reasoning upon semantically bound information \cite{Kanerva:Hyper_dym11, PlateThesis}.

The fundamental difference between distributed and localist representations of data is as follows: 
in traditional (localist) computing architectures each bit and its position within a structure of bits are significant (for example, a field in a database has a predefined offset amongst other fields and a symbolic value has unique representation in ASCII codes); whereas in a distributed representation all entities are represented by binary random vectors of very high dimension. These representations are also called Binary Spatter Codes (BSC), though for the remainder of the article we use term \textit{HD-vector} when referring to BSC codes, for reasons of brevity. 
{\it High dimensionality} refers to that fact that in HD-vectors, several thousand positions (of binary numbers) are used for representing a single entity; \cite{Kanerva:Hyper_dym11} proposes the use of vectors of 10000 binary elements.  
Such entities have the following useful properties:

\subsubsection{Randomness}

Randomness means that the values on each position of an HD-vector are independent of each other, and "0" and "1" components are equally probable. In very high dimensions, the distances from any arbitrary chosen HD-vector to more than 99.99 \%  of all other vectors in the representation space are concentrated around  0.5 Hamming distance. Interested readers are referred to \cite{Kanerva:Hyper_dym11} and \cite{KanervaBook} for comprehensive analysis of probabilistic properties of the hyperdimensional representation space.

Denote the density of a randomly generated HD-vector (i.e.\
the number of ones in a HD-vector) as $k$. The probability of
picking a random vector of length $d$ with density $k$, where
the probability of 1's appearance equals $p$ is described by described by the binomial distribution (\ref{eq:bindistr}):

\begin{equation}
\label{eq:bindistr} 
\text{Pr}(k,d,p)= \binom{d}{k}p^{k}(1-p)^{d-k}.
\end{equation}

When $d$ is in the range of several thousand binary elements, the  calculation of the
binomial coefficient requires sophisticated computations. The following equations use an
approximation of the binomial distribution, via the de Moivre-Laplace theorem:

\begin{equation}
\label{eq:bindistrappr} 
\text{Pr}(k,d,p)\approx \displaystyle\frac{e^{\frac{-(k-dp)^2}{2dp(1-p)}}}{\sqrt{2\pi dp(1-p)}}
\end{equation}

\subsubsection{Similarity metric}

A similarity between two binary representation is characterized by Hamming distance, which (for two vectors) measures the number of positions in which they differ: 

\begin{center}
$\Delta_H(A,B)= \frac{\lVert  A \otimes B \rVert_{1}}{d} = \frac{\sum^{d-1}_{i=0}a_i \otimes b_i}{d}$,
\end{center}

where $a_i$, $b_i$ are bits on positions $i$ in vectors $A$ and $B$ of dimension $d$, and where $\otimes$ denotes the bit-wise XOR operation.

\subsubsection{Generation of HD-vectors}

Random binary vectors with the above properties can be generated with Zadoff-Chu sequences \cite{Zadoff-Chu}, a method widely used in telecommunications to generate pseudo-orthogonal preambles. Using this approach a sequence of $K$ vectors, which are pseudo-orthogonal  to a given initial random HD-vector $A$ (i.e.\ Hamming distance between them approximately equals 0.5) are obtained by cyclically shifting $A$ by $i$ positions, where $1< i \leq K \leq d$. Further in the article this operation is denoted as $\text{Sh}(A,i)$. The cyclic shift operation has the following properties: 

\begin{itemize}
  \item it is invertible, i.e. if $B=\text{Sh}(A,i)$ then $A = \text{Sh}(B,-i)$;
  \item it is associative in the sense that $\text{Sh}(B,i+j)=\text{Sh}(\text{Sh}(B,i),j)=\text{Sh}(\text{Sh}(B,j),i)$;
  \item it preserves Hamming weight of the result:  \\ $\lVert  B \rVert_{1}= \lVert  \text{Sh}(B,i) \rVert_{1}$; and
  \item the result is dissimilar to the vector being shifted: \\ $\frac{\lVert  B \otimes \text{Sh}(B,i) \rVert_{1}}{d} \approx 0.5$.
\end{itemize}

Note that the cyclic shift is a special case of the permutation operation \cite{Kanerva:Hyper_dym11}. In the context of VSA, permutations were previously used to encode sequences of semantically bound elements.

\subsubsection{Bundling of vectors}

Joining several entities into one structure is done by the bundling operation; it is implemented by a thresholded sum of the HD-vectors representing the entities. A bit-wise thresholded sum of $n$ vectors results in 0  when $n/2$ or more arguments are 0, and 1 otherwise. Furthermore terms "thresholded sum" and "majority sum" are used interchangeably and denoted as $[A+B+C]$. The relevant  properties of the majority sum are: 

\begin{itemize}
  \item the result is a random vector, i.e. the number of `1' components is approximately equal to the number of `0' components;
  \item the result is similar to all vectors included in the sum;
  \item the number of vectors involved into MAJORITY sum must be odd; 
  \item the more HD-vectors that are involved in a majority operation, the closer the Hamming distance between the resultant vector and any HD-vector component is to 0.5; and
  \item if several copies of any vector are included into a majority sum, the resultant vector is closer to the dominating vector than to other components.      
\end{itemize}
 
 The algebra on VSA includes other operations e.g., binding, permutation \cite{Kanerva:Hyper_dym11}. Since we do not use them in this article, we omit their definitions and properties.

\section{Holographic Graph Neuron}
\label{sect:hologen}
This section presents one of the main contributions of this paper: the adoption of the VSA data representation for implementation of the HGN approach.

\subsection{Motivation for Holographic Graph Neuron and outline of the solution}

\begin{figure}[t!]
\centering
\includegraphics[width=\columnwidth]{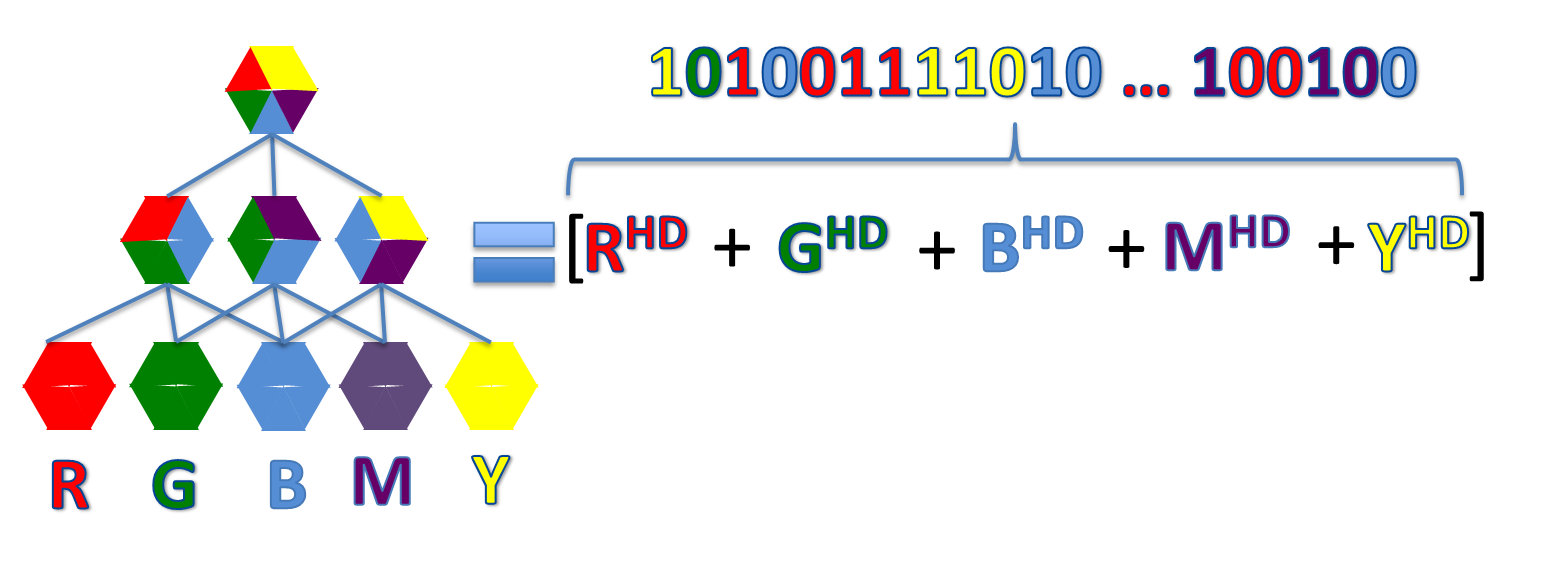}
\caption{A High level illustration of the proposed solution: representation of
the Hierarchical Graph Neuron using Vector Symbolic Architecture. 
} 
\label{fig:illustration}
\end{figure}

An important issue in hierarchical models is the overhead of resource requirements, specifically 
with regards to the number of processing elements required. 
We propose a holographic approach, which enables a flat
GN array to operate with higher level of accuracy and comparable recall time than that of HGN, 
without the need for a complex topology and additional nodes. The high-level
logic of the proposed solution is illustrated in Figure \ref{fig:illustration}.
Essentially, we replace the need for maintaining topological relationships by
compacting parts of the pattern observed by GNs into one
holographic representation.

\subsection{Encoding}

In the case of HoloGN, all its elements (i.e. symbols of
individual neurons) are indexed uniquely and the index of a particular element
is derived as a function of the GN's ID.
Let $IV_j$ be an {\it initialization high-dimensional vector} for GN $j$.
The initialized vectors for different GNs are chosen to be mutually
orthogonal. Then the HD-index of element $i$ in GN $j$ is computed as
$E^{HD}_{(j,i)}=\text{Sh}(IV_j, i)$, where $\text{Sh}()$ is a cyclic shift operation resulting in
the generation of a vector orthogonal to $IV_j$ HD-vector \cite{Zadoff-Chu, BICA14}.
 
\subsection{Construction of VSA-representation of activated GNs}

Let $n$ be the number of individual Graph Neurons. When a GN array is exposed to a pattern, the activated elements communicate
their HD-represented indices to all other GNs; the holographic
representation of the activated elements is then:

\begin{equation}
\label{eq:hgn}
\text{HGN}=  [  \sum\limits_{j=1}^{n}(E^{HD}_j) ],
\end{equation}

where $E^{HD}_j$ is the HD-index of the activated element in $text{GN}_j$, and the addition operation is the bundling operation, or thresholded sum as described in
\ref{sect:hologr_struct}.  As discussed previously, in the resultant HD-vector the
Hamming distances between each component and the composition vector is strictly less
than 0.5. We will utilize this property later when constructing data structures for recall of patterns in HoloGN.

\subsection{Data structures for storing and retrieving holographic 
representations in HGN elements}
HoloGN will store the holographic representation of the entire pattern
(\ref{eq:hgn})  observed across all GNs. A possible architecture is for 
all memorized
patterns to be collected and stored centrally at a processing node. 

Depending on the particular application of HoloGN, the memorized patterns could be
stored either in an unsorted list or in bundles. The first mode of storing HoloGN patterns corresponds
to the case where the structure of the observed patterns is unknown; the
latter mode is used in the case of a {\it supervised
learning}. The next section describes different HoloGN usage and recall strategies.

\section{HoloGN recall strategies }

This section introduces and evaluates the performance of two major
recall modes: the one-shot case and the case of supervised learning. 
 
 \subsection{Time-efficient $\xi$-accurate recall in an unsorted HoloGN storage }
 
 The common steps in both recall modes is the procedure for the time-efficient search over an
 (unsorted) list of HoloGN records. 
 Recall that all manipulations with
 VSA-encoded entities are done using simple bit-wise arithmetic operations as
 well, and calculations to obtain the Hamming distance between entities. However, for this article we assume no particular optimized implementation of VSA's bit-wise operations; this is because such operations are tailored to the architectures of 
 specific microprocessors, which operate with words of 
 substantially lower dimensionalities (typically 32 or 64 bits). Therefore, adopting these
 methods for implementing the bit-wise operations on words of thousands of bits
 would be cumbersome. Instead, an easily analyzable  computational model is adopted in this article, which could also be adapted to an implementation on specialized computing architectures.
 
 In what follows, each HoloGN pattern $\textbf{h}_i$ is modelled as a row vector of
  $d$ elements. The list of stored HoloGN patterns
 is therefore modelled as an $l \times d$ matrix, where $l$ is the number
 of the learned (stored) HoloGN patterns. Denote this matrix as $\textbf{H}$. The task
of recalling a pattern  with a target accuracy of $\xi$ ($\xi<0.5$) is formulated
as finding the rows $\textbf{h}_i$ in $\textbf{H}$ with Hamming distance to the query pattern $\textbf{h}_q$ less than or equal to $\xi$. 

The conventional  way of computing Hamming distance between vectors would be to perform the
following sequence of computations for each row in $\textbf{H}$: 1) perform an element-wise 
XOR with the vector query; 2) sum up all elements in the intermediate
result; and then 3) divide the result by the dimensionality of the vectors.
The performance of the two implementations of this method using Matlab's
$repmat$ and $bxfun$ functions are shown by the top two curves in Figure
\ref{fig:decoding}. The curves demonstrate linear but rapid increase in the
recall time with the increase in the number of the stored patterns. The lowest
curve in the figure shows the performance of matrix-vector multiplication of the
same size, which is chosen as the reference case. The results were obtained
 on a Intel Core i7-3520M 2.9 GHz  machine with  Windows 7 operating system
 using one processor.

\subsubsection{Binary spatter codes as complex numbers} In order to improve  the
efficiency of calculating Hamming distances over  a vast number of HoloGN
patterns, we propose representing HoloGN patterns using complex numbers, 
where a binary $0$ would be represented by $\sqrt{-1}$ (i.e. the complex number) and a binary
$1$ would remain $1$. The intuition behind this transformation is simple: 
multiplication of bits in the same position should produce three  outcomes:
$-1=j \times j$, $1=1 \times 1$ and $j=1\times j$. That is, the multiplication of
 two similar bits would produce a real number and the multiplication of two
different bits produces a imaginary number. In this way the sum  of the imaginary 
parts over all positions in the resulting vector, divided by dimensionality $d$, will correspond to the Hamming distance between the two vectors. 
Thus, the suggested method allows us to implement the calculation of Hamming distance through the standard method of matrix-vector
multiplication, as illustrated below:

\begin{equation}
\small
\begin{matrix}
\textbf{H}\times \textbf{h}_q=
\begin{pmatrix}
  \sqrt{-1} & 1 & \cdots & \sqrt{-1} \\
  1 & 1 & \cdots & \sqrt{-1} \\
  \vdots  & \vdots  & \ddots & \vdots  \\
  \sqrt{-1} & \sqrt{-1} & \cdots & 1
\end{pmatrix} 
\times 
 \begin{pmatrix} 
  \sqrt{-1}  \\
  1  \\
  \vdots   \\
  \sqrt{-1}  
\end{pmatrix} =
\\
~\\
=
 \begin{pmatrix}
  254+ 1633j\\
  617+ 3824j\\
  \vdots   \\
  548+ 4952j
\end{pmatrix}. 
\end{matrix}
\end{equation}
 
\begin{figure}[htb]
\centering
\includegraphics[width=\columnwidth]{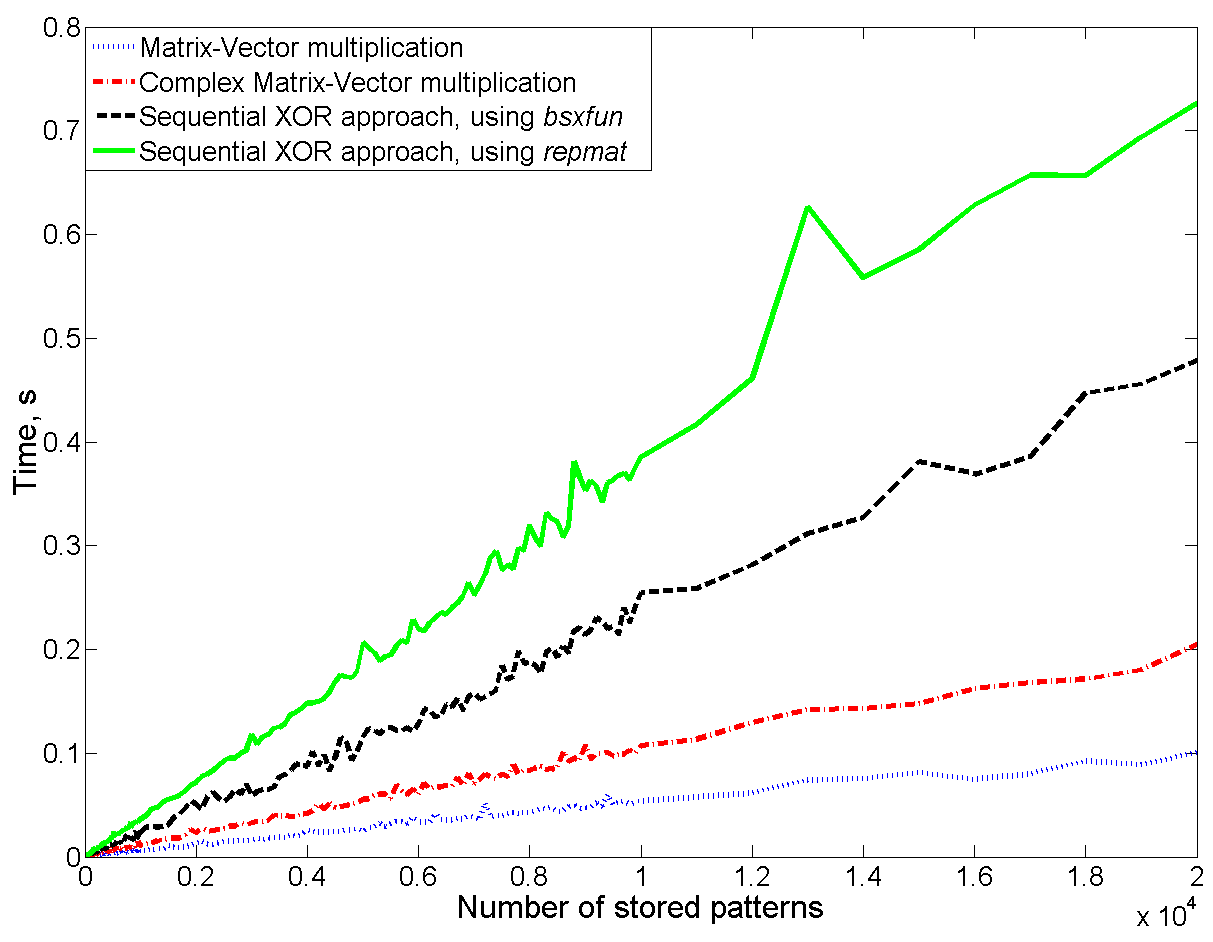}
\caption{A comparison of the different approaches to recalling patterns, showing the time taken to calculate the Hamming distance against the number of previously presented patterns.}
\label{fig:decoding}
\end{figure}

 The performance of the proposed method is illustrated by the red dashed curve
 in Figure \ref{fig:decoding}.  It demonstrates that the calculation of the Hamming
 distance is only two times slower than the usual matrix-vector
 multiplication. Specifically, to compute Hamming distance from the target vector to each of 20 000 stored
 patterns takes approximately 200 ms on the test machine. Further
 optimization of the matrix multiplication and execution on parallel
 architectures hint on realistic bounds on the recall time over
 extremely large numbers of patterns.

\subsection{Case-study 1: Best-match probing under one-shot learning }
The first usage of HoloGN is probing the existence of the target
query pattern  amongst the previously memorized patterns. The perfect match in
this case would be indicated by a Hamming distance of zero. The deviation
from zero, therefore, reflects the degree of proximity of the query 
to one or several stored HoloGN patterns. In the following example the accuracy
of the HoloGN recall was compared to the performance of the original HGN approach. For the sake of fair comparison the 7 by 5 pixels letters
of the Roman alphabet (as in \cite{10_NasutionKhan}) were used. 

\begin{figure}[htb]
\centering
\includegraphics[width=1.0\columnwidth]{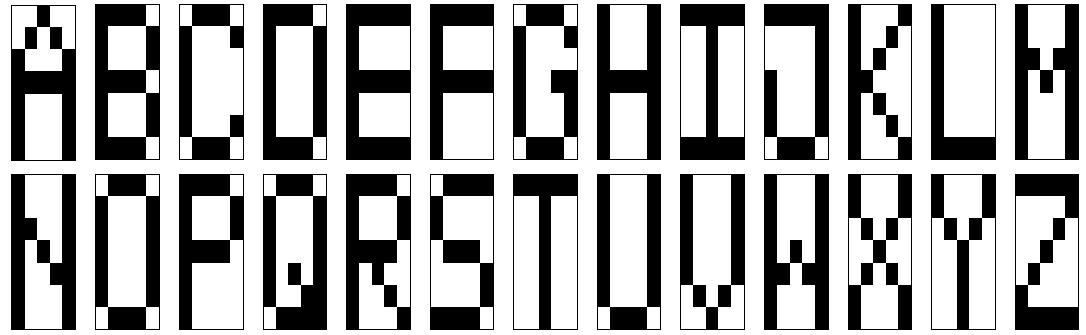}
\caption{List of letters images for comparison.}
\label{fig:letters}
\end{figure}

\begin{figure}[htb]
\centering
\includegraphics[width=1.0\columnwidth]{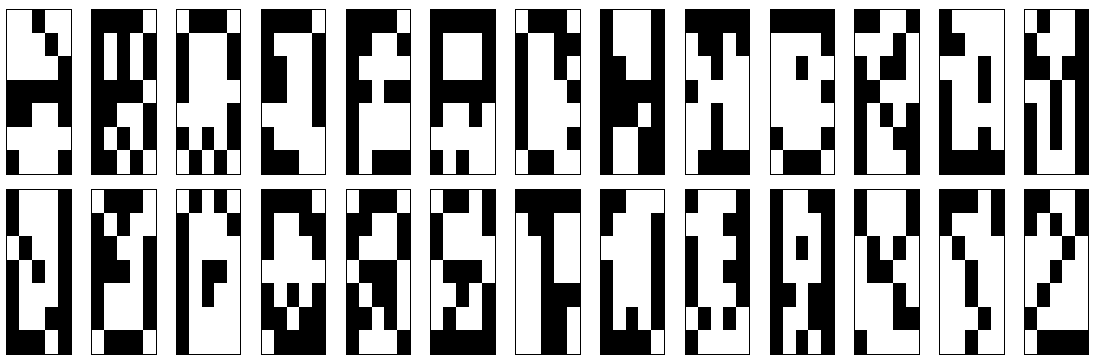}
\caption{An example of presented for recall images distorted by 14.3\%.}
\label{fig:letters_noisy}
\end{figure}

In the memorizing phase a set of noise-free images of letters illustrated in
Figure \ref{fig:letters} was presented to both architectures. In the recall phase
images of the same letters distorted with different levels of random distortions (between 1
bit corresponding to a distortion of 2.9\% of the pattern's size and 5 bits
equivalent to 14.3\% distortion) were presented to the
architectures for the recall.  An example of a noisy input is presented in
Figure \ref{fig:letters_noisy}. In the case of  HoloGN the pattern with the
lowest Hamming distance to the presented distorted pattern was returned as the output.

\begin{figure*}[htb]
\centering
\includegraphics[width=2.0\columnwidth]{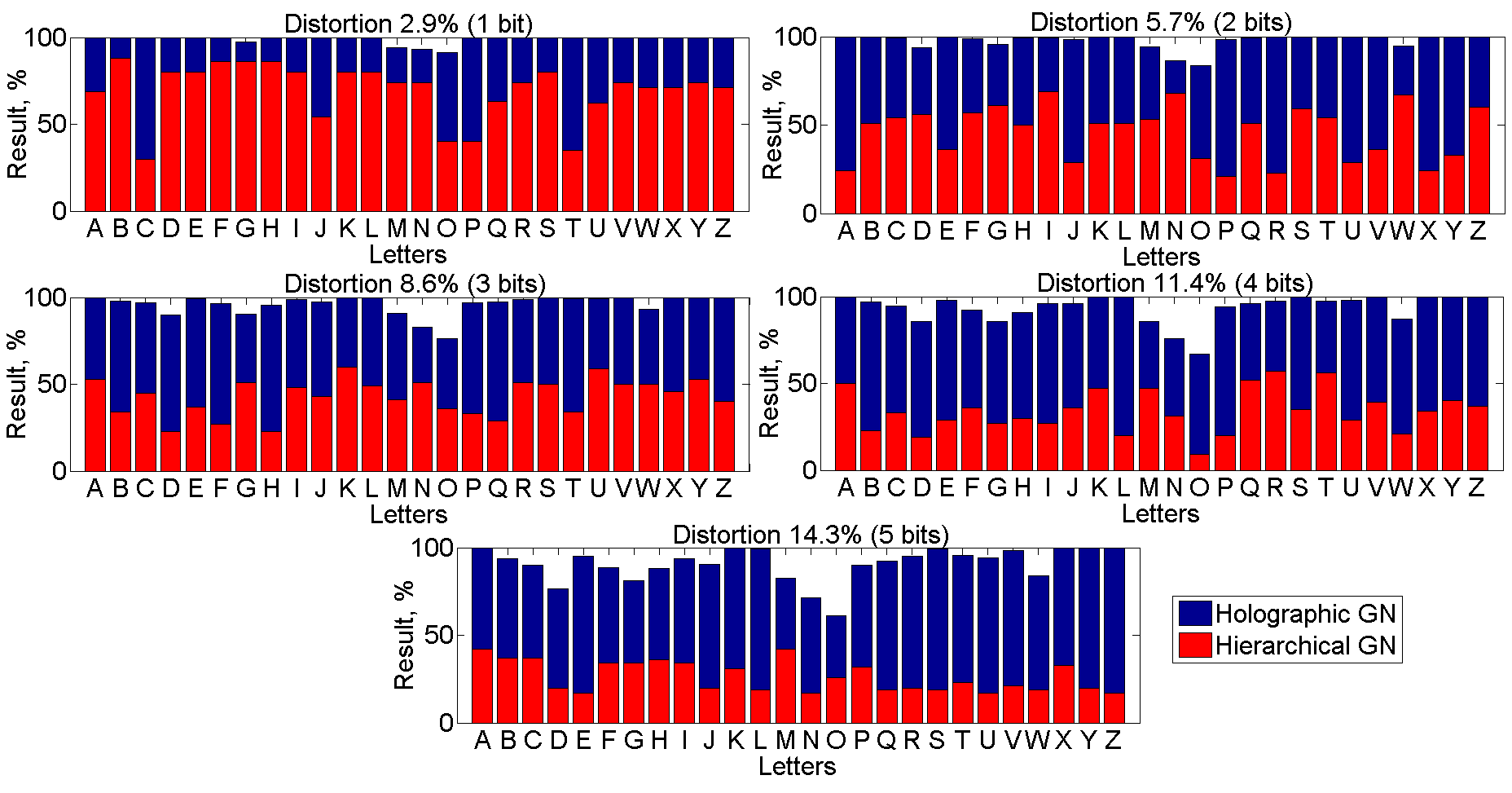}
\caption{Results from testing black and white images of letters using recall patterns with distortions ranging from 2.9 \% to 14.3 \% .}
\label{fig:noisecomp}
\end{figure*}

Figure \ref{fig:noisecomp}  presents the results of the accuracy
comparison between the recall results for the HoloGN approach and the reference HGN architecture.  To obtain the results 1000
distorted images of each letter for every level of distortion were presented
for recall. The charts show the percentage of the correct
recall output.  The analysis shows that the performance of the HoloGN-based associative
memory at least matches that of the original approach. In most of cases 
HoloGN appears to be more accurate when recalling patterns with high level
of distortion.

 \subsection{Case-study 2: HoloGN recall under the supervised learning }
The analysis presented above is a very positive result for the proposed bio-inspired associative memory based pattern processing
architecture, since the accuracy of the original HGN
approach was demonstrated to be as accurate as artificial neural
networks with back-propagation \cite{10_NasutionKhan}. While establishing formal
relationships to the framework of artificial neural networksis
outside the scope of this work, this section presents the results of the pattern
recognition accuracy of the HoloGN architecture under the supervised learning. In this case the HoloGN is presented with a series of randomly distorted
patterns for each letter with different level of distortion (between 3\% and 43\%) as
exemplified in Figure \ref{fig:letters_noisy}. In the experiments up to 500
patterns for each letter and every level of distortion were presented for
memorizing. For the particular level of distortion $i$ all $e$ presented patterns of
the particular letter $L_i$ were bundled to a single HoloGN representation as

\begin{equation}
\textbf{h}(L)=[ \sum\limits_{i=1}^{e}( \text{HGN}(L_i)) ].
\end{equation}

Thus by the end of the learning phase the HoloGN list will contain 26 high
dimensional bundles, each jointly representing all (presented) distorted variants
of the particular letter. In the recall phase for each distortion level  HoloGN
was presented with 500 new distorted patterns of each letter. The accuracy of
the recall was measured as the percentage of the correctly recognized letters averaged over the alphabet. 

\begin{figure}[htb]
\centering
\includegraphics[width=1.0\columnwidth]{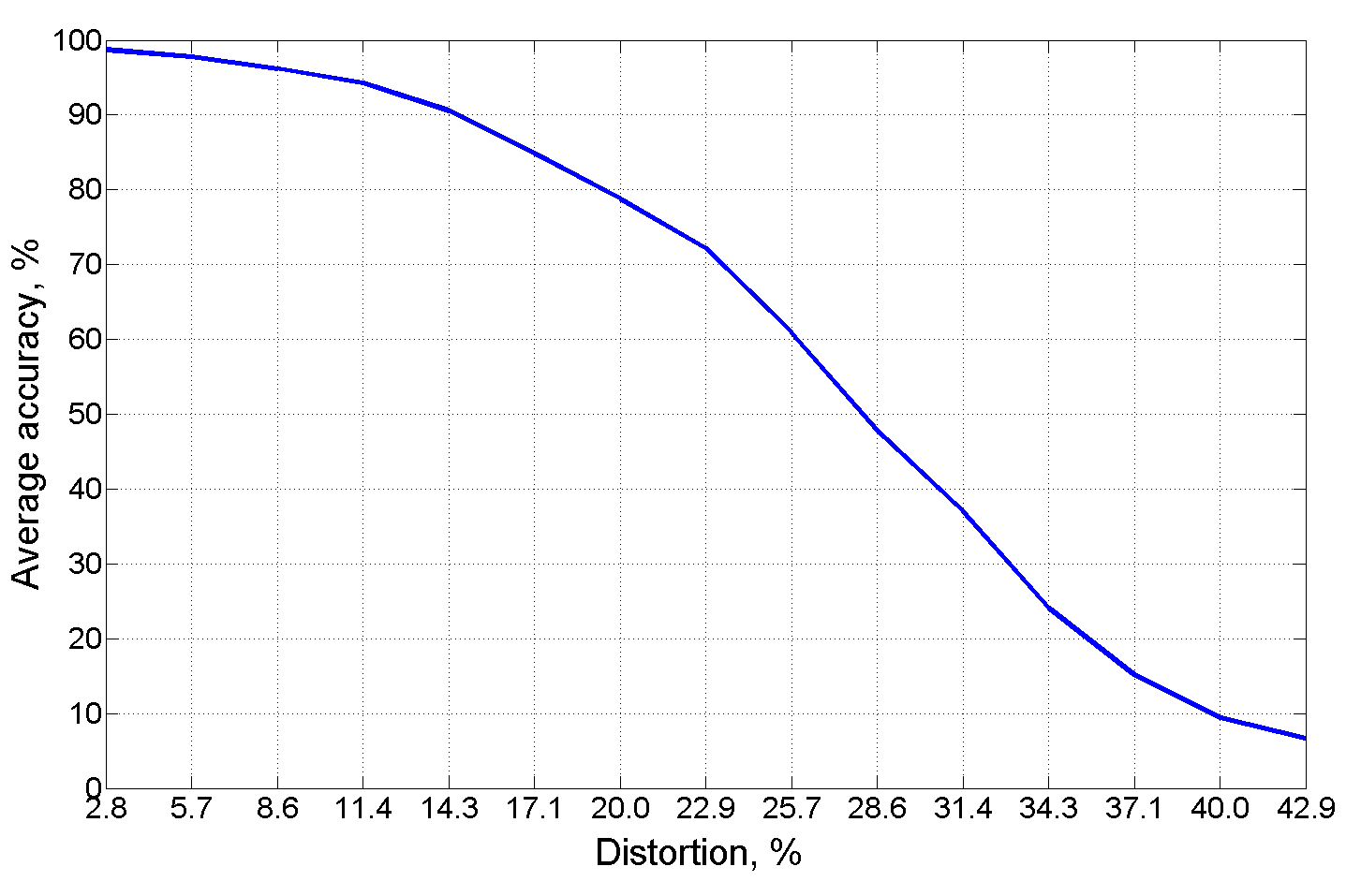}
\caption{Average accuracy of HoloGN recall under the supervised learning memorization as
a function of the distortion level.}
\label{fig:supervized_learning_distortion}
\end{figure}

Figure \ref{fig:supervized_learning_distortion} illustrates the obtained
results: 90\% accurate recall was observed when learning symbols distorted up to
15\%. While the accuracy predictably decreases rapidly with the increase of distortion in
the presented patterns, a reasonable 80\% recall accuracy was
observed for learning sets with 7 bits distortion (20\%).

\begin{figure}[htb]
\centering
\includegraphics[width=1.0\columnwidth]{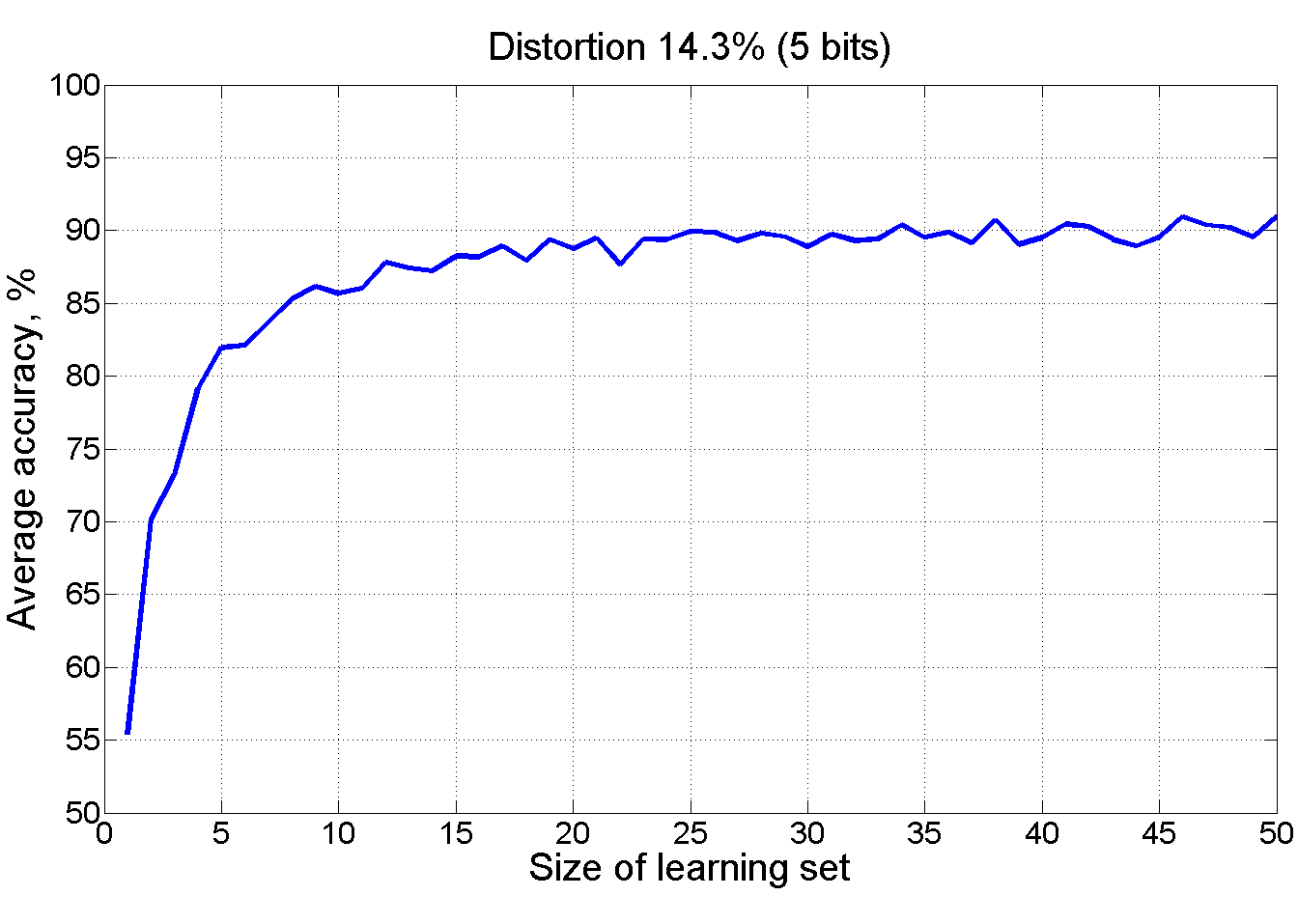}
\caption{Accuracy of HoloGN under the supervised learning memorization as a
function of the number of presented examples for a given level of distortion.}
\label{fig:supervized_learning_examples}
\end{figure}
 
Figure \ref{fig:supervized_learning_examples} illustrates the
convergence of the HoloGN recall accuracy with the number of presented
noisy samples for the case of 14.3\% distortion (5 bits). This is a
definitely positive result for the presented architecture, which illustrates the
suitability of the HoloGN in applications requiring supervised learning.

\section{Pattern decoding and subpattern-based analysis}
\label{sect:vectorperf}

There is a class of pattern recognition applications which requires
an understanding of the details of the recall results. For example, when a recall returns several possible patterns of given recall accuracy, the task would be to understand 
the overlapping elements.
This section considers two aspects of this task: a robust decoding of elementary components out of a distributed VSA representation; and a quantitative metric of the similarity via direct comparison of distributed representations, without the need for decoding those representations.

The VSA approach of representing data structures by definition makes decoding of the individual components a  tedious task, requiring a brute force test on the inclusion of
all possible high-dimensional codewords for each GN. The majority sum, which is used for creating HoloGN representations of the observed patterns imposes a limit
 on the number high-dimensional codewords operands, above which a robust decoding of the individual operands is impossible. 
 
\subsection{Preliminaries}

Denote the density of a randomly generated HD-vector (i.e. the number of {\it
ones} in a HD-vector) as $k$. The probability of picking a random
vector of length $d$ with density $k$, where the probability of 1's appearance, defined as $p$, is 
described by (\ref{eq:bindistrappr}).  The mean density of a random vector is equal to $d \cdot p$. Note that in reality the density of
randomly generated HD-vectors will obviously deviate from the mean value.   However,
according to (\ref{eq:bindistr}) density $k$ is approaches the mean value with
the increase of dimensionality $d$.
In other, words the probability of generating  HD-vector with $k>>d \cdot p$ or
$k<<d \cdot p$ decreases with the increase of dimensionality $d$. 

\begin{figure}[htb]% [t!]
\centering
\includegraphics[width=1.0\columnwidth]{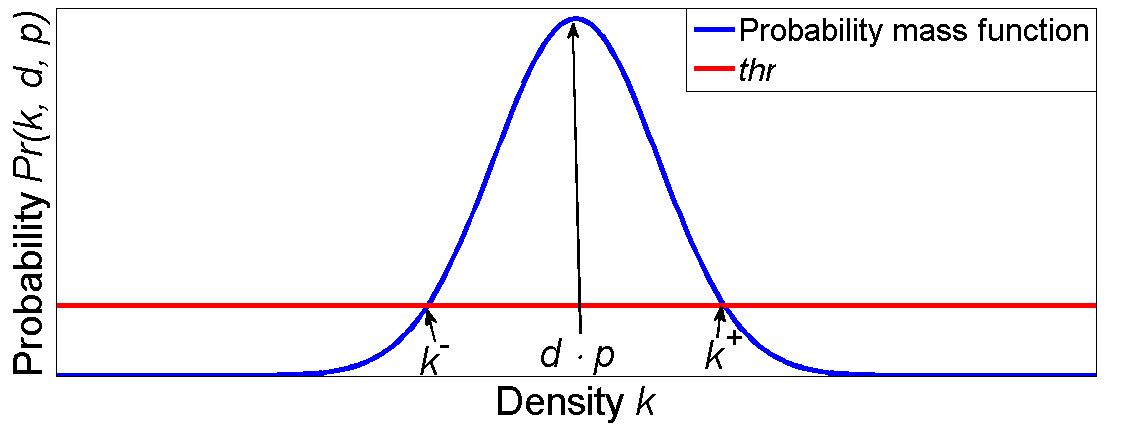}
\caption{Binomial distribution and its parameters describing HD-vectors.}
\label{fig:bindist}
\end{figure}

Define $thr$ as the threshold probability of generating a vector with a certain deviation of density being negligibly small.
Let $k^{-}$ and  $k^{+}$ characterize the lower and the upper bounds of the interval of possible densities. This is illustrated in Figure \ref{fig:bindist}. The bounds for a given $d$, $p$ and $thr$ are calculated using (\ref{eq:bindistrappr}). The bounds are calculated according to (\ref{eq:binthrlow}) and (\ref{eq:binthrup}). The value of threshold is chosen to be small ($10^{-6}$).

\begin{equation}
\label{eq:binthrlow} 
k^{-}(d,p,thr)=\max_{k}(\text{Pr}(k,d,p)<=thr|k<(d \cdot p))
\end{equation}

\begin{equation}
\label{eq:binthrup} 
k^{+}(d,p,thr)=\min_{k}(\text{Pr}(k,d,p)<=thr|k>(d \cdot p))
\end{equation}

\subsection{Capacity of HoloGN representations}
\label{sect:capacity}

Suppose there exists an item memory \cite{Kanerva:Hyper_dym11} containing HD-vectors representing atomic
concepts\footnote{In the case of HoloGN an atomic concept is the code for the
particular HoloGN element.}. Recall that when several
HD-vectors are bundled by the majority sum the noise of flipped bits increases with the number of components. For a given dimensionality $d$ there is a limit on the number of
bundled HD-vectors beyond which the resulting HD-vector becomes orthogonal to every component vector; hence the A {\it capacity} of the resulting vector is defined as the maximal number of mutually orthogonal HD-vectors
which can be  robustly decoded from their majority sum composition by probing the item memory.

In order to characterize the capacity of the composition vector for a given
dimensionality, one needs to characterize the level of noise $p_n$  introduced by the
bundling operation. This is calculated as in \cite{Kanerva97} by (\ref{eq:majority}), where $n$ is a number of atomic vectors in the resulting majority sum vector.

\begin{equation}
\label{eq:majority}
p_n(n)= \frac{ \frac{1}{2} -\begin{pmatrix} n-1 \\ 0.5 \cdot (n-1)  \end{pmatrix}}{2^n}.
\end{equation}

\begin{figure}[htb]% [t!]
\centering
\includegraphics[width=1.0\columnwidth]{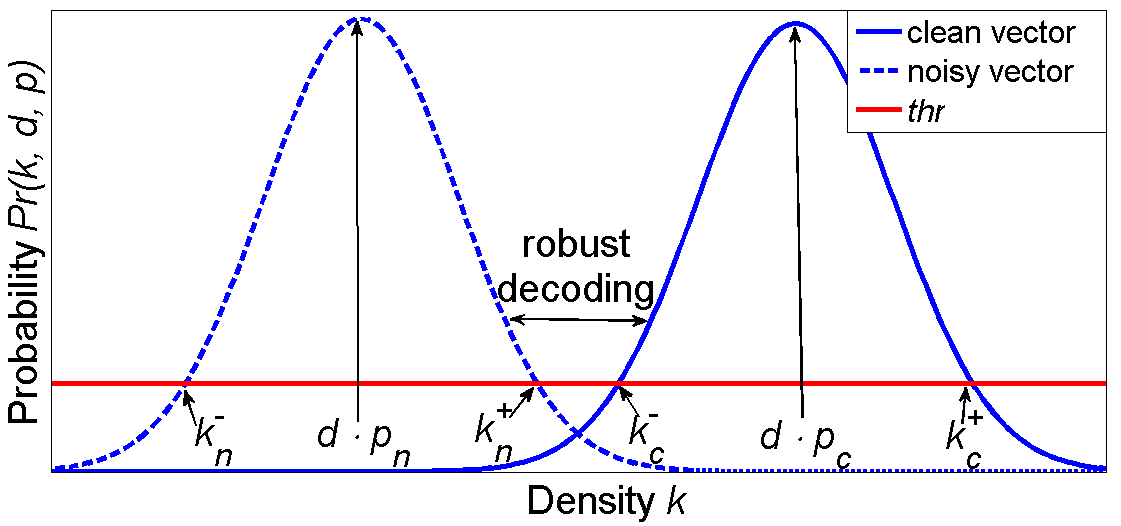}
\caption{Explanation of vector's capacity. Solid line represents random HD-vector. Dashed line corresponds to noise introduced by majority sum. }
\label{fig:bindist_cap}
\end{figure}

\begin{figure}[htb]% [t!]
\centering
\includegraphics[width=1.0\columnwidth]{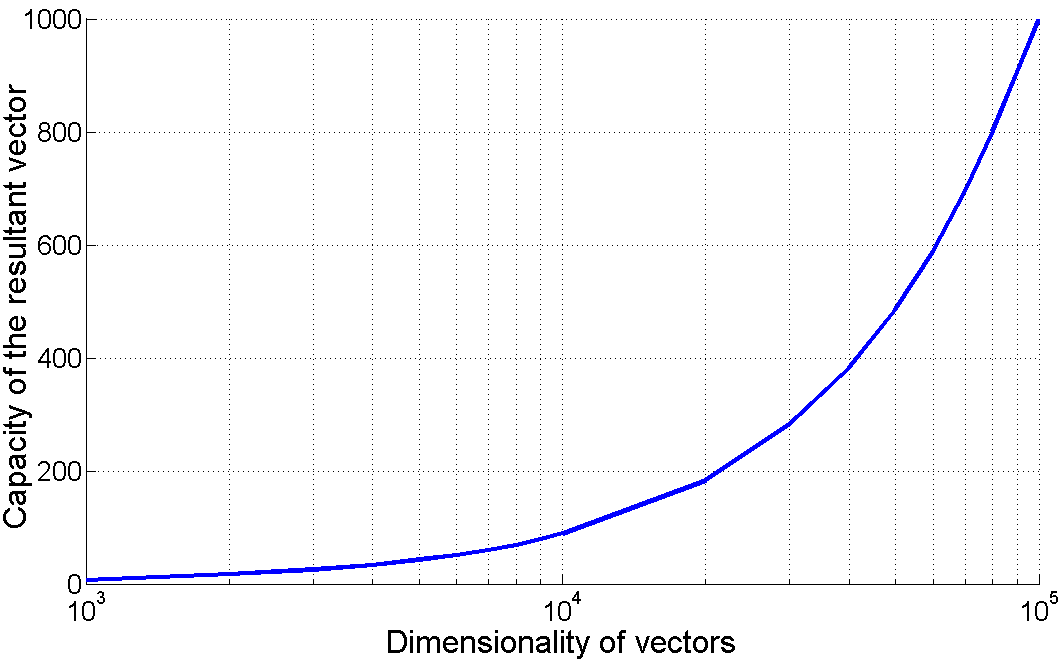}
\caption{Capacity of the component vector versus the dimensionality. The calculations use a the threshold of $thr=10^{-6}$}
\label{fig:capacity}
\end{figure}

Consider an arbitrary HD-vector $\textbf{A}$ to be decoded from a majority sum composition. Let $\textbf{N}$ be a vector of noise imposed by the majority sum operation. Since the components are mutually orthogonal, the density of ones in the noise vector is also described the binomial distribution $\text{Pr}(k,d,p_n)$.  As each new vector is added the level of noise increases, hence the mean of the noise vector density will approach 0.5 as illustrated in Figure \ref{fig:capacity}. Due to the properties of high dimensional space, vector $\textbf{A}$ will be undecodable when the upper bound $k^{+}(d,p_n,thr)$ of the density of noise vector $\textbf{N}$ approaches the lower bound $k^{-}(d,0.5,thr)$. That is, the noisy version of $\textbf{A}$ becomes orthogonal to its clean version. This logic is illustrated in Figure \ref{fig:bindist_cap}, where
the resulting majority sum vector is orthogonal to all components. Thus the capacity of the distributed representation with dimensionality $d$ is computed by:

\begin{equation}
\label{eq:binthr}
\begin{split}
&\text{Capacity}(d,thr)= \\
&=\max_{n}(k^{+}(d,p_n(n),thr) \leq k^{-}(d,0.5,thr))
\end{split}
\end{equation}

Figure \ref{fig:capacity}  presents the capacity of HD-vector of different dimensionalities calculated for threshold probability $thr=10^{-6}$.  Specifically for $d=10000$ bits the capacity of the robustly decodable VSA is 89 vectors.

\subsection{Calculation of the number of common component vectors in two resulting vectors}
\label{sect:mpn}

Given the rather conservative limits on the number of robustly decodable elements in a distributed representation it is important that the proposed HoloGN architecture can estimate the similarity between different patterns \textit{without} decoding them. This subsection provides a method for the quantitatively measuring the number of overlapping elements as a function of their relative Hamming distance. Denote $m$ and $n$ as lengths of two patterns, $m<=n$, and denote $c$ as the number of common elements in these patterns. Let \textbf{M} be a $c \times d$ matrix of common elements where each row contains a random HD-vector of dimension $d$ encoding element $c_i$. Denote an arbitrary column of matrix \textbf{M} as \textbf{C}. Since rows in \textbf{M} are independent, the density of \textbf{ones} in each column also follows the binomial distribution with $p=0.5$ and length $c$. Denote the number of \textbf{ones} in column \textbf{C} as $||\textbf{C}||_1$.
 
In order to calculate the Hamming distance between the distributed representations of two patterns with known $m$, $n$ and $c$,  consider all possible cases when bits in the same position are different. The Hamming distance between two patterns can be estimated with equation (\ref{eq:probkmn}):

\begin{equation}
\label{eq:probkmn}
\begin{split}
 \Delta_H= p(c,m,n)= & \displaystyle\sum_{||\textbf{C}||_1=0}^{c}\displaystyle  \frac{\binom{c}{||\textbf{C}||_1}}{2^c} \cdot (p_{1}(m,c,||\textbf{C}||_1) \cdot \\ \cdot p_{0}(n,c,||\textbf{C}||_1)+
&p_{0}(m,c,||\textbf{C}||_1)\cdot p_{1}(n,c,||\textbf{C}||_1));  
\end{split}
\end{equation}

where $p_{i}(j,c,||\textbf{C}||_1)$ stands for the probability of having  $i$ (0 or 1), when the representation consists of $j=m$ or $j=n$ atomic vectors and $c$ of these vectors are overlapped. 

Due to the symmetry in the calculation of probabilities, $p_{i}(j,c,||\textbf{C}||_1)$ is presented only for the case of $p_{1}(m,c,||\textbf{C}||_1)$. There are three possible cases for calculation of $p_{1}(m,c,||\textbf{C}||_1)$: 
\begin{itemize}
  \item if $||\textbf{C}||_1$ is more than $m/2$, then the result of the majority sum is `1', i.e. $p_1$ is 1;
  \item if number of possible `1's is less than $m/2$, then probability of $p_1$ is 0;
  \item otherwise the probability should take into account all possible combinations, and their probabilities. 
\end{itemize}

Equation (\ref{eq:probm1}) specifies probability for  $p_{1}(m,c,||\textbf{C}||_1)$, and equation (\ref{eq:probm0}) does the same for $p_{0}(m,c,||\textbf{C}||_1)$.

\begin{equation}
\label{eq:probm1} 
p_{1}(m,c,||\textbf{C}||_1)=\begin{cases} 1 ,  \mbox{  when } ||\textbf{C}||_1>\frac{m}{2}  \\
0 ,  \mbox{  when } (m-c) < \frac{m+1}{2}-||\textbf{C}||_1 \\
\displaystyle\frac{\displaystyle\sum_{i=(\frac{m+1}{2}-||\textbf{C}||_1)}^{(m-c)}\binom{m-c}{i} }{2^{(m-c)}}  ,  \mbox{  otherwise }
\end{cases}
\end{equation}

\begin{equation}
\label{eq:probm0} 
p_{0} (m,c,||\textbf{C}||_1)=\begin{cases} 1 ,  \mbox{  when } (c-||\textbf{C}||_1)>\frac{m}{2}  \\
0 ,  \mbox{  when } (m-c) < \frac{m+1}{2}-(c-||\textbf{C}||_1) \\
\displaystyle\frac{\displaystyle\sum_{i=(\frac{m+1}{2}-(c-||\textbf{C}||_1))}^{(m-c)}\binom{m-c}{i} }{2^{(m-c)}}, \mbox{  otherwise }
\end{cases}
\end{equation}

\begin{figure}[htb]
\centering
\includegraphics[width=1.0\columnwidth]{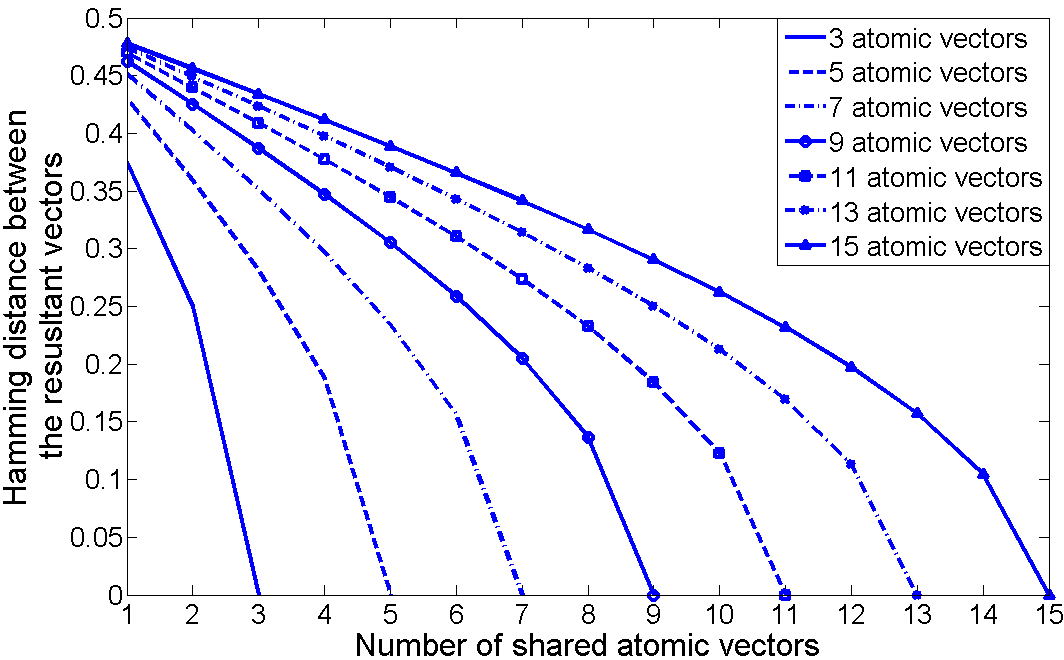}
\caption{The Hamming distance between two resulting vectors against number of components in common. The number of atomic vectors is the same, $m=n$.}
\label{fig:mnk}
\end{figure}

Figure \ref{fig:mnk} shows the Hamming distances between two resulting vectors for different numbers of overlapping vectors. The results show that the larger the number of common elements, the smaller the Hamming distance between resulting vectors.

This method opens a way towards constructing and analyzing patterns far beyond VSA's robustly decodable capacity. The problem with practical application of this method, however, comes with the rapid convergence of the Hamming distance indicator to 0.5, making the difference between analyzable patterns indistinguishable as illustrated in Figure \ref{fig:mnk}. For example, for HoloGN representations of patterns with 15 elements, sub-patterns of 3 overlapped elements are robustly detected, while patterns with fewer overlapped elements are indistinguishable. Thus, the minimal number of overlapped elements in two patterns which can be robustly detected using the Hamming distance indicator is called bundle's \textit{sensitivity}.

The analysis of the sensitivity is similar to the analysis of the capacity of VSA representation in section \ref{sect:capacity}. For two patterns of length $m$ and $n$ elements, and $c$ overlapped components, the sensitivity is calculated by (\ref{eq:sens–two}):

\begin{equation}
\label{eq:sens–two}
\begin{split}
&\text{Sensitivity}(d,thr,m,n)= \\
&\min_{c}(k^{+}(d,p(c,n,m),thr)<=k^{-}(d,0.5,thr))
\end{split}
\end{equation}

\begin{figure}[htb]
\centering
\includegraphics[width=1.0\columnwidth]{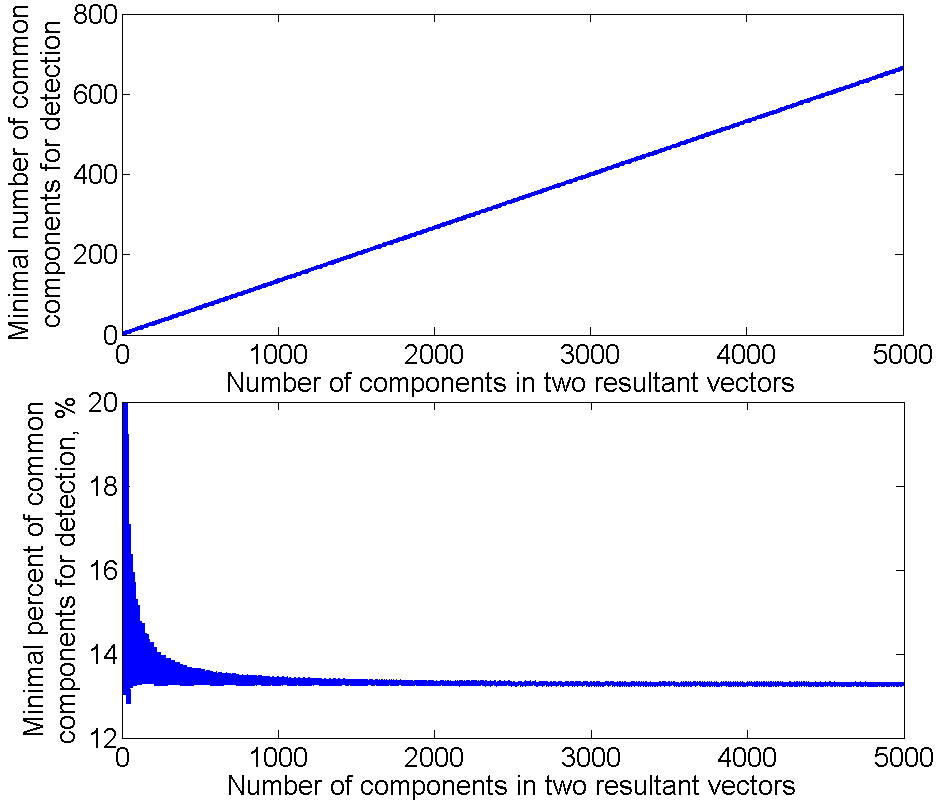}
\caption{Minimal number of common components, which is sensible between two patterns of the same size against the size of patterns, $d=10000$, $thr=10^{-6}$.}
\label{fig:senstwo}
\end{figure}

Figure \ref{fig:senstwo} demonstrates  the development of the sensitivity threshold with the number of elements in the compared patterns. The results show that the number of components for robust detection grows linearly with the size of pattern. Patterns with more than 500 elements should contain at least 14 \% overlapped elements  to be robustly detected by the proposed method. 

\section{Conclusion}
\label{sect:conclusions}
This article presented Holographic Graph Neuron - a novel approach for memorizing 
patterns of generic sensor stimuli. HoloGN is built upon the previous Graph
Neuron algorithm and adopts  a Vector Symbolic representation for encoding of
the Graph Neuron's states. The adoption of the Vector Symbolic representation 
ensures a one-layered design for the approach, which implies the computational
simplicity of the operations. The presented approach possesses the number of
uniques properties. Prior to all, it enables a linear (with respect to the
number of stored entries) time search for an arbitrary sub-pattern. While
maintaining the previously reported properties of
Hierarchical Graph Neuron, HoloGN is also improving the noise  resistance of the
architecture by this substantially improving the accuracy of pattern recall.

%\bibliography{references}
%\bibliography{references.bbl}

\end{document}